\pdfoutput=1
\documentclass[11pt]{article}
\usepackage[final]{coling}
\usepackage{times}
\usepackage{latexsym}
\usepackage[T1]{fontenc}
\usepackage[utf8]{inputenc}
\usepackage{microtype}
\usepackage{inconsolata}
\usepackage{times}
\usepackage{latexsym}
\usepackage{amssymb}
\usepackage{amsmath}
\usepackage{amsthm}
\usepackage{booktabs}
\usepackage{enumitem}
\usepackage{graphicx}
\usepackage{color}
\usepackage{multirow}
\usepackage{bbding}
\usepackage{tabularx}

\newcommand{\modelname}{{Comateformer}}

\title{\modelname: Combined Attention Transformer for Semantic Sentence Matching}

\author{Bo Li $^{1,2}$ \thanks{The corresponding author.}, Di Liang$^{2}$ and  Zixin Zhang $^{1,2}$ \\
$^1$School of Software, Tsinghua University, Beijing, China  \\
$^2$Baidu Inc., Beijing, China\\}

\begin{document}
\maketitle
\begin{abstract}
The Transformer-based model have made significant strides in semantic matching tasks by capturing connections between phrase pairs.
However, to assess the relevance of sentence pairs, it is insufficient to just examine the general similarity between the sentences. 
It is crucial to also consider the tiny subtleties that differentiate them from each other. Regrettably, attention softmax operations in transformers tend to miss these subtle differences. 
To this end, in this work, we propose a novel semantic sentence matching model named \textbf{\textit{Com}}bined \textbf{\textit{Atte}}ntion Network based on Trans\textbf{\textit{former}} model (\textbf{\modelname}). In \modelname~model, we design a novel transformer-based quasi-attention mechanism with compositional properties.
Unlike traditional attention mechanisms that merely adjust the weights of input tokens, our proposed method learns how to combine, subtract, or resize specific vectors when building a representation. 
Moreover, our proposed approach builds on the intuition of similarity and dissimilarity (negative affinity) when calculating dual affinity scores. 
This allows for a more meaningful representation of relationships between sentences. 
To evaluate the performance of our proposed model, we conducted extensive experiments on ten public real-world datasets and robustness testing. 
Experimental results show that our method achieves consistent improvements.
\end{abstract}

\section{Introduction}
\label{sec:intro}

Semantic sentence matching (SSM) is a core method used in the field of natural language processing (NLP) with the objective of comparing and discerning the semantic correlation between two given phrases.
In paraphrase identification \cite{madnani-etal-2012-examining}, SSM is used to determine whether two sentences are paraphrase or not.
In natural language inference task \cite{bowman2015large} also known as recognizing textual entailment, SSM determines whether a hypothesis sentence can reasonably be inferred from a given premise sentence.
In the answer sentence selection task \cite{wang2020multi}, SSM is employed to assess the relevance between query-answer pairs and rank all candidate answers.
In large language models such as GPT \cite{ouyang2022training} and LLaMA \cite{touvron2023llama}, SSM can be used for parallel corpus alignment and data denoising.
However, the task of establishing the logical and semantic link between two statements is not straightforward, mostly because of the challenge posed by the semantic gap \cite{Im_Cho_2017}.

\begin{figure}[t]
\centering
\includegraphics[width=0.9\linewidth]{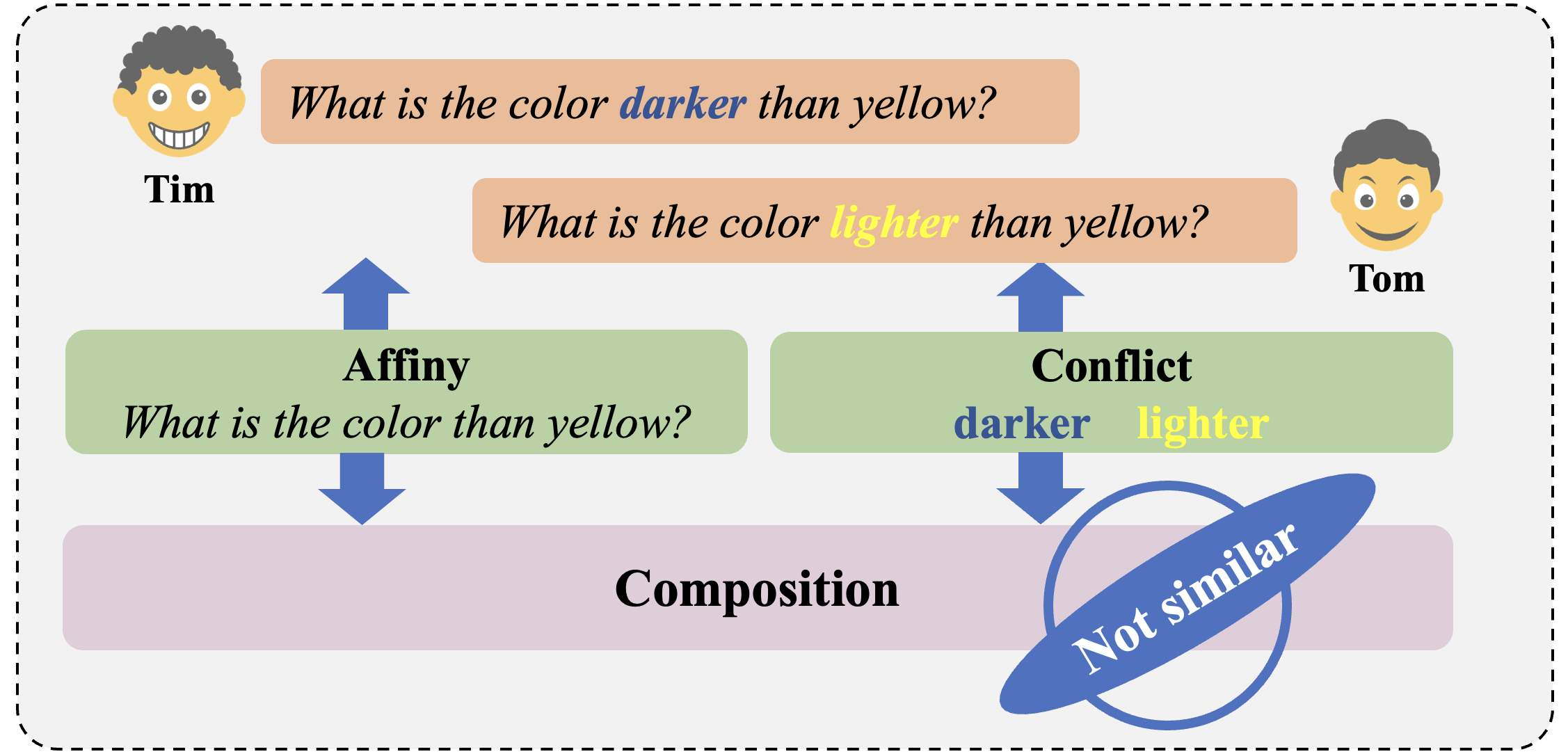}
\caption{The Combined Attention Network Example for Semantic Sentence Matching.}
\label{fig:example} 
\end{figure}

Across the rich history of semantic sentence matching research, there have been two main streams of studies for solving this problem.
The first is representation based method which encodes each of the sentences and obtain their representation vectors in low-dimensional latent space and then utilize the parameterized matching function for the final matching scores, which focuses on how to get good sentence representations \cite{reimers2019sentence,wang2020multi,xue2023dual}. 
Another type of semantic matching model is interaction-based, which directly align the sentences based on the attention mechanism and aggregate the matching score to directly make the final decision which focuses on how to align the word pairs in the sentence pair \cite{chen2016enhanced,liang2019adaptive,song-etal-2022-improving-semantic,liang2019asynchronous,wang2022dabert}.
Recently large-scale pre-trained language models such as BERT \cite{devlin2018bert}, RoBERTa \cite{liu2019roberta}, are becoming more popular in multiple NLP tasks. Because of their high efficiency and effectiveness in contextual information modeling and sentence level encoding, pre-trained models are also wildly used in semantic matching tasks and achieve significant improvement.Recent work attempts to integrate external knowledge  \cite{zhang2020semantics,xia2021using,bai2021syntax} into PLMs. Meanwhile, leveraging external knowledge to enhance PLMs has been proven to be highly useful for multiple NLP tasks \cite{kiperwasser2018scheduled}. Recent work also attempts to enhance the performance of BERT by injecting knowledge into it, such as SemBERT \cite{zhang2020semantics}, UER-BERT \cite{xia2021using}, Syntax-BERT \cite{bai2021syntax}, DABERT \cite{wang2022dabert} and so on.

Although previous studies have provided some insights, those models (e.g., BERT, RoBERTa) do not perform well in distinguishing sentence pairs with high literal similarities but different semantics. Figure \ref{fig:example} exemplifies an instance that is afflicted by this issue.
Although the sentence pairs in this figure are semantically different, they are too similar in literal for those pre-trained language models to distinguish accurately. 
One significant factor is that while the model possesses the capability to assess the level of similarity in overall semantics, it fails to account for the nuanced distinctions present within individual texts. Because for text pairs with highly similar matching words, the overall semantic difference is often caused by different local differences. 
Furthermore, an obvious feature of the attention model is that it can learn relative importance, that is, assign different weights to input values, and the Softmax operator is its core. Softmax makes the weight of core words higher and the weight of non-core words lower. 
Sparsegen \cite{martins2016softmax} have proved that equipping with attention mechanism with more flexible structure, models can generate more powerful representations. In this paper, we also focus on enhancing the attention mechanism in transformer-based pre-trained models to better integrate difference information between sentence pairs.
We hypothesize that paying more attention to the fine-grained semantic differences, explicitly modeling the difference and affinity vectors together will further improve the performance of pre-trained model. 

In this paper, we propose a novel approach named \modelname, designed exclusively for semantic matching tasks. It replaces vanilla attention in the transformer with combined attention, which works similarly to vanilla attention, but with several key fundamental differences. 
First, instead of learning relative importance (a weighted sum), combinatorial attention learns combinations of tokens that decide whether to add, subtract, or scale inputs. In other words, our method removes softmax operation because it deviates from the original motivation of attention, so we refer to our method as combined attention. 
Second, we introduce a quadratically scaled attention matrix, ultimately learning a multiplicative combination of similarity and dissimilarity. We hypothesize that a more flexible design can lead to more expressive and robust models, leading to better performance.
To achieve this, we propose two modules to implement the above description. The first is the dual-affinity module, which introduces a negative affinity matrix $N$ in addition to the original affinity matrix $E$, and the affinity matrix $E$ is obtained by labeling the attention formula, that is, $e_{ij}$ = $a_i$ $\cdot$ $b_j$. 
In contrast to $E$, the negative affinity matrix $N$ learns a dissimilarity metric ($n_{ij}$ = $a_i$ - $b_j$) for modeling the differences between word pairs.
Subsequently, we introduce a combination mechanism that combines $tanh(E)$ and $sigmoid(N)$ to form a quasi-attention matrix $M$. In this case, the first term $tanh(E)$ controls the addition and subtraction of vectors, while the auxiliary Affinity $N$ can be interpreted as a gating mechanism that scales unnecessary features when needed.
We conduct a series of experiments on 10 datasets and the experimental results show that the method achieves consistent improvements.

The main contributions of this paper are as follows.
\begin{itemize}
\item First, we conduct a comprehensive study of the subtle differences between sentence pairs and propose a new method named \modelname~for semantic matching tasks, which has two distinct kinds of functions to represent the interaction between phrase pairs from various viewpoints,and the softmax function was eliminated from the attention mechanism, resulting in an increased receptive field and enhanced capacity to catch tiny differences. 

\item Secondly, we explicitly integrate \modelname~into both pre-trained and non-pre-trained models, and the results showed that the proposed method can provide greater expressive power, and it can fully discover the inherent complex relationships between sentence pairs for effective semantic matching.

\item Finally, we carry out a series of experiments on 10 matching datasets and robustness testing datasets.  Experimental results show that \modelname~has achieved consistent improvements, especially in the robustness test, achieving an average improvement of 5\% over BERT.
The effectiveness of \modelname~is further supported by a case study and an attention distribution analysis, which illustrate the model's nuanced handling of sentence pair interactions and its ability to focus on both commonalities and differences within the text.
\end{itemize}

\begin{figure*}[t]
\centering
\includegraphics[width=0.8\textwidth]{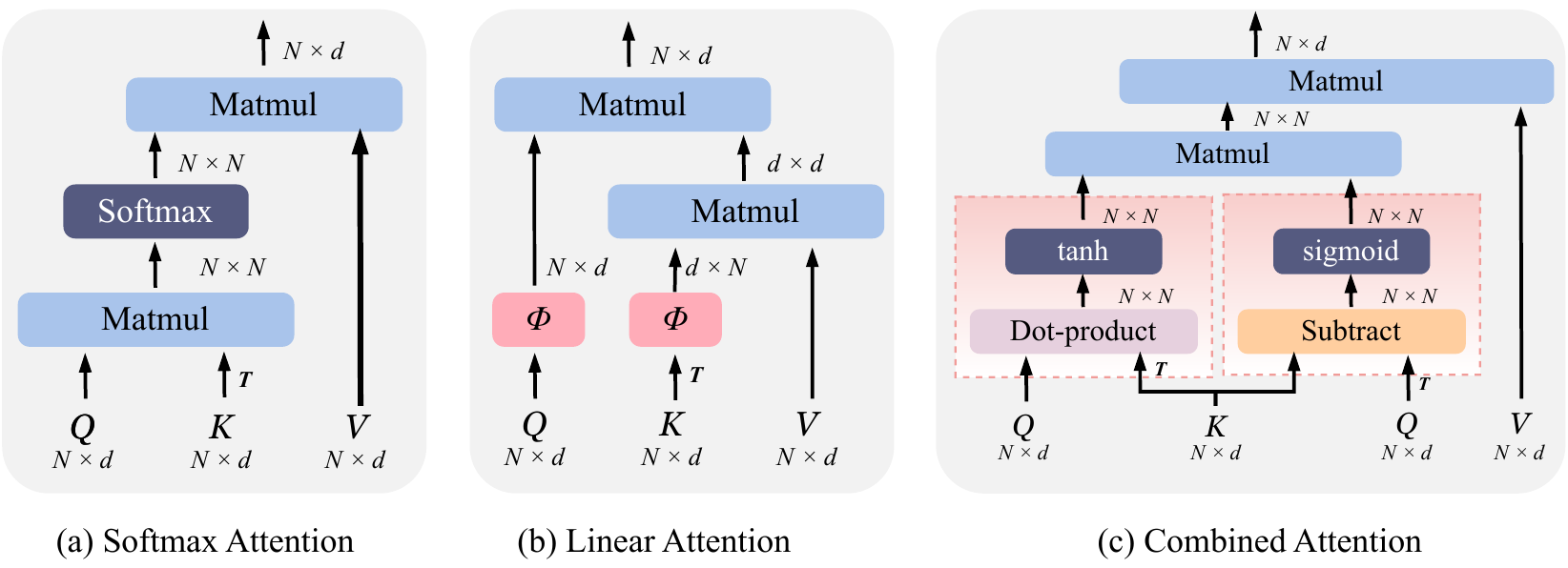}
\caption{Difference between Softmax attention, Linear attention and Combined attention. Softmax attention computes the similarity between all Q-K pairs. Linear attention applies mapping function $\Phi$(·) to Q and K respectively. Our Combined Attention models both global affinity and local difference information, thus achieving dual perception of affinity and non-affinity, with higher fine-grained differentiation advantages.
}
\label{fig:module2}
\end{figure*}

\section{Related work}

\label{sec:related_works}

Our work relates to several work in the literature: Semantic Sentence Matching, Robustness test. We will discuss each of these as follows.
\subsection{Semantic Sentence Matching}
SSM is a focal point within the field of NLP, witnessing significant advancements over the years.
It mainly fell into two categories: traditional neural network based methods and pre-trained language model based methods. 

\paragraph{Traditional Neural Network based methods.}
Early approaches to SSM were predominantly reliant on traditional methods such as syntactic features, transformations, and relation extraction~\cite{romano2006investigating,wang2007jeopardy}. 
These methods, while effective for specific tasks, were inherently limited in their scope and generalizability.
With the advent of large-scale annotated datasets~\cite{bowman2015large,zheng2022robust} and the proliferation of deep learning algorithms, neural network models have made great progress in SSM. 
The incorporation of attention mechanisms marked a pivotal shift, offering richer information for sentence matching by elucidating alignment and dependency relationships between sentences~\cite{conneau2017supervised,choi2018learning,gui2018transferring}. 
These mechanisms endowed models with the ability to capture nuanced semantic similarities beyond the lexical surface.
Concurrently, joint methods that leveraged cross-features through attention mechanisms were introduced to address the limitations of sentence-encoding methods, enhancing performance by capturing word- or phrase-level alignments~\cite{wang2017bilateral,gong2017natural}. The architectural advancements, including the use of residual connections, facilitated the stable increase of network depth, preserving information from lower layers~\cite{he2016deep}.
\cite{liu2016learning,peng2020enhanced} emphasis on the sequential information and the semantic interdependence of sequences. \cite{xu2020enhanced,liu2023time} used distinct convolutional filters to capture the local context. By supplying alignment and dependence relationships between two sentences, the well-established attention processes provided greater information for sentence matching. This was accomplished by giving the information. \cite{cho2015describing} used an attention method to extract the salient components inside sentences, record the semantic connections, and appropriately align the pieces of two phrases. \cite{liu2023local,ma2022searching} employed a stacked multi-layer Bi-LSTM with Alignment Factorization to quantify the various levels of features between two texts. Convolutional Neural Network (CNN)  focus on the local context extraction with different kernels, and Recurrent Neural Networks (RNN) are mainly utilized to capture the sequential information and semantic dependency. \cite{yang2019simple} utilized a multi-layer encoding technique and fusion block based on a CNN structure to construct a rapid and highly effective phrase matching model. \cite{dong2020distilling,fei2022cqg} utilized GNN to leverage the structural information of input sentences in order to achieve full sentence connection modeling. 

\paragraph{Pre-trained Language Model based methods.}
Recently, the pre-trained language models, most notably BERT~\cite{devlin2018bert}, revolutionized SSM by providing powerful sentence representations through self-supervised learning on vast corpora. This paradigm shift allowed for transfer learning across various NLP tasks, significantly accelerating research progress.
One way to enhance the performance of pre-trained models is by modifying the input encoding and utilizing self-supervised pre-trained tasks. 
XLNet~\cite{yang2019xlnet} utilized a recently developed PLM task to reduce the disparity between pre-trained tasks and subsequent tasks. 
Moreover, there are other noteworthy advancements in this field, including RoBERTa~\cite{liu2019roberta} and CharBERT~\cite{ma2020charbert}. \cite{liang2019asynchronous,chen2016enhanced} utilizes cross-features as an attention module to express the word-level or phrase-level alignments for performance improvements, and aggregates these integrated information to acquire similarity. DenseNet~\cite{xue2024question} belongs to the joint approaches which utilizes densely-connected recurrent and co-attentive information to enhance representation. 
Meanwhile, there is a trend to utilize explicit NLP knowledge to improve sentence representation~\cite{liu2024resolving}. For example,~\cite{tymoshenko-moschitti-2018-cross,liu2018structured} used the syntactic dependencies to enhance the sentence representations. 
The NLP knowledge-enhanced matching models have also adapted to the interaction-based models. 
For example, MIX~\cite{li2024local} utilizes POS and named-entity tags as prior features.
SemBERT \cite{zhang2020semantics} concatenates semantic role annotation to enhance BERT. 
UERBERT \cite{xia2021using} chooses to inject synonym knowledge. 
SyntaxBERT \cite{bai2021syntax} integrates the syntax tree into transformer-based models. 

The above work has achieved significant advancements in sentence semantic matching, which has motivated us to maximize the utilization of sophisticated neural networks and pre-trained techniques for sentence semantic modeling. However, these models possess the capability to assess the level of similarity in overall semantics, it fails to account for the nuanced distinctions present within individual texts. Because for text pairs with highly similar matching words, the overall semantic difference is often caused by different local differences.
 
\subsection{Robustness Test}
Although neural network models have achieved human-like or even superior results in multiple tasks, they still face the insufficient robustness problem in real application scenarios \cite{gui2021textflint}. Tiny literal changes may cause misjudgments. Especially in some cases where fine-grained semantic needs to be discriminated. Besides, most of the current work utilizes one single metric to evaluate their model, may overestimate model capability and lack a fine-grained assessment of model robustness. 
Therefore, recent work starts to focus on robustness research from multiple perspectives. TextFlint incorporates multiple transformations to provide comprehensive robustness analysis. \cite{li2021searching} provide an overall benchmark for current work on adversarial attacks. And \cite{liu2021explainaboard} propose a more comprehensive evaluation system and add more detailed output analysis indicators.

\section{Method}
\label{sec:method}

\subsection{Task Definition}
\label{sec:task definition}
In a formal manner, it is possible to describe each instance of sentence pairings as a triple (Q, P, y). Here, Q represents a phrase of length N, denoted as ($q_{1}$, ..., $q_{N}$), P represents another sentence of length M, denoted as ($p_{1}$, ..., $p_{M}$), and y $\in$ Y is the label representing the relation between Q and P. In the job of identifying paraphrases, Q and P represent two sentences. The variable y is used to denote the outcome, where Y can take the values of either 0 or 1. Specifically, y = 1 indicates that Q and P are paraphrases of each other, whereas y = 0 indicates that they are not paraphrases. In the context of a natural language inference task, the premise sentence is denoted as Q, the hypothesis sentence as P, and the variable y represents the possible outcomes of the task, namely inference, contradiction, or neutral. Inference refers to the situation where P can be logically deduced from Q, contradiction indicates that P cannot be a valid condition for Q, and neutral signifies that P and Q are unrelated to each other.

\begin{figure*}[t]
\centering
\includegraphics[width=0.8\textwidth]{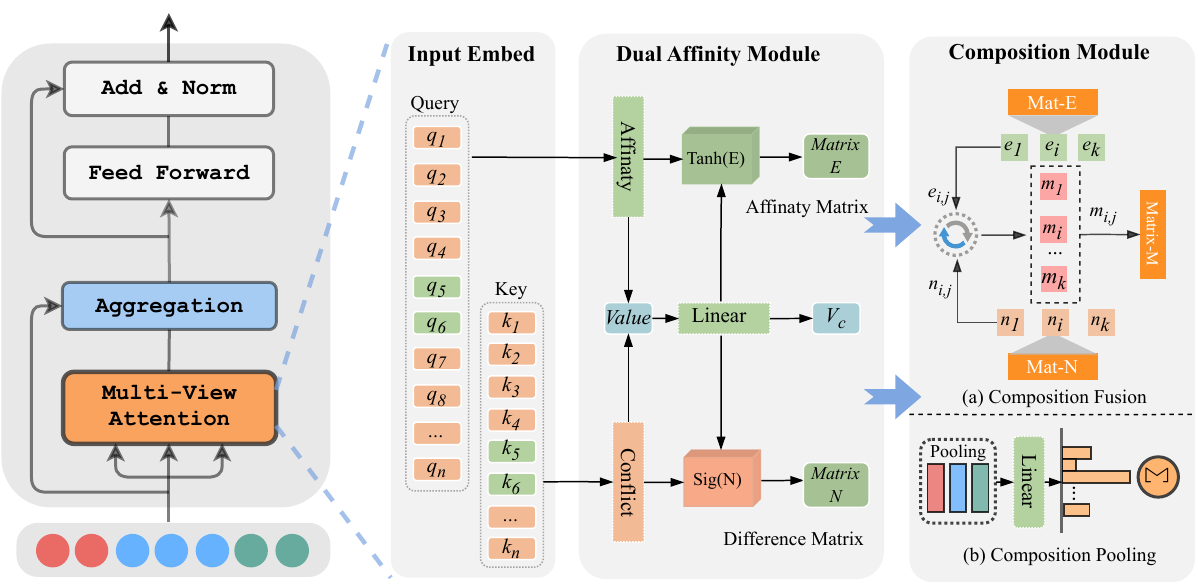}
\caption{The overall architecture of incorporating \modelname~to transformer Model.}
\label{fig:module} 
\end{figure*}

A comparison between \textbf{\textit{\modelname~}} and classical attention is included in Figure \ref{fig:module2}.
It consists of two parts under the combined attention framework. First, we model the interaction of sentence pairs from different perspectives using two different types of functions. Next, we removed the softmax operation in attention, and gave the attention a wider receptive field and a more subtle difference capture ability. 
Two sentences are input as $\mathbf{A} \in R^{N_a \times d}$ and 
$\mathbf{B} \in R^{N_b \times d}$, Where $N_a, N_b$ is the length of sequences A and B. They are padded to the same length $N$ by default. And d is the dimension of the input vector, and returns a combination with the same dimension express. Note that the input is generic as it can be applied to interactive attention for dual sequences and self-attention for single sequences. In the case of single-sequence attention, the variables A and B typically denote identical sequences.

\subsection{Dual Affinity Module}
In this module, we design two different functions, affinity function and difference function, to compare the affinity and difference of vectors between two sentences. First, we compute the pairwise affinities between each word in A and B via the dot product:

\begin{equation}
    E_{ij} = \alpha \times F_E(a_i)F_E(b_j).
\end{equation}

This function computes the pairwise similarity between any two elements in A and B. In this procedure, $F_E$(.) represents a parameterized function, such as a standard linear/nonlinear function. 
Additionally, $\alpha$ represents a scaling constant and a non-negative hyperparameter, which can be thought of as a temperature setting that adjusts saturation.
Next, as a measure pairwise of negativity (i.e., dissimilarity) between each word in A and B, we perform the following calculation:

\begin{equation}
    N_{ij} = \beta \times ||F_N(a_i) - F_N(b_j)||.
\end{equation}

In this function, we introduce a parameterized function $F_N$(.) and a scaling constant $\beta$, while preserving the L1-Norm \textit{$l_1$}. It is noteworthy to mention that in practice we can make parameters shared between $F_E(.)$ and $F_N(.)$. Meanwhile, the affinity matrix N has the same dimensions as the affinity matrix \textit{$E$}. Our argument posits that capturing features of different properties (e.g. subtractive compositionality) in attention models is crucial for semantic matching tasks. The fundamental concept underlying negative distance involves utilizing negative affinity values as a gating mechanism to represent negative qualities, a capability that is absent in the original attention method.

\subsection{Compositional Attention Module}
In the typical vanilla attention, \textit{softmax} is the core component, which is applied to the matrix $E$ to normalize it. Hence, multiplying the normalization matrix of E with the original input sequence yields a vanilla attention pooled representation (aligned representation), where each element in sentence A pools all relevant information for all elements in sentence B. The combined attention we propose is completely different from vanilla attention. First, it has no softmax operation. Specifically, we use the following equations for attention modeling:

\begin{equation}
    \mathbf{M} = \tanh(\textit{E}) \displaystyle \odot sigmoid(N),
\end{equation}
 
where M is the final attention matrix in the combined attention mechanism, which is an element-wise multiplication between two matrices.

\paragraph{Normalization of matrix N.} Since $N$ is constructed from negative L1 distances, it is clear that \textit{sigmoid(N)} $\in$ [0, 0.5]. Therefore, to ensure that \textit{sigmoid(N)} lies in the range [0, 1], we center the matrix $N$ so that its mean is zero:

\begin{equation}
    \mathbf{N} = \mathit{N} - \textit{Mean(N)}.
\end{equation}

Intuitively, by scaling the matrix $N$, we preserve the ability to scale up and down the median of the \textit{tanh(E)} matrix, since \textit{sigmoid(N)} has a saturation region between 0 and 1, so it behaves more like a gating mechanism. At the same time we also try the second form of scaling, as an alternative to centering:

\begin{equation}
    \mathbf{M} = \tanh(\textit{E}) \displaystyle \odot (2* sigmoid(N)).
\end{equation}

Empirically, we have found that this approach is also very effective.

\paragraph{Temperature.} We introduced the hyperparameters $\alpha, \beta$ that control the size of \textit{E} and \textit{N} in the previous sections. Intuitively, these hyperparameters control and affect the temperature of the tanh and sigmoid functions. In other words, high values of $\alpha, \beta$ will enforce hard-form combined pooling. 
In this task we set $\alpha = 1$ and $\beta$ = 1.

Finally, we apply the Compositional Attention Matrix $M$ to the input sequences A and B with the following formula:

\begin{equation}
    \hat{A} = M \times B \quad and \quad \hat{B} = M^T \times A.
\end{equation}
    
And the two sentences are update as $\hat{A} \in R^{N_a \times d}$ and 
$\hat{B} \in R^{N_b \times d}$.  
Taking $\hat{A}$ as an example, each element $A_i$ in A traverses sentence B and determines whether it contains the token in sentence B by adding (+1), subtracting (-1) or deleting (×0). Similarly, each element in sentence B traverses sentence A and decides to add, subtract, or delete a token from A.
Intuitively, which can capture both affinity and dissimilarity features, facilitating rich and expressive representations, unlike typical attention pooling methods that operate on sequences.

\subsection{Incorporating \modelname~to Transformer}
 
As shown in Figure \ref{fig:module}, which shows the location of \modelname~integrated in the transformer and the schematic diagram of the specific modules of \modelname.  The original Transformers \cite{vaswani2017attention} employ a self-attention mechanism, which can be interpreted as cross-attention on the same sequence. Our \modelname~replaces the original attention module with de-softmaxed Dual Affinity Module, that is, the original Transformer internal attention equation $A = softmax(\frac{Q K^T}{\sqrt{d_k}}) * V$ is now changed to:

\begin{equation}
    \mathbf{A} = \tanh(\frac{\mathbf{Q K}^T}{\sqrt{d_k}}) \displaystyle \odot sigmoid(\frac{\mathbf{G(Q K})}{\sqrt{d_k}}) * \mathbf{V},
\end{equation}

where G(.) is the negation of outer L1 distance between all rows of Q against all rows of K. 
We either apply centering to ($\frac{G(Q K)}{\sqrt{d_k}}*V$) or
2 * $sigmoid(\frac{G(Q K)}{\sqrt{d_k}})*V$ to ensure the value is in
[0, 1]. Finally, both affinity matrices are learned by transforming Q, K, V only once. 

\subsection{Incorporating \modelname~to PLMs}
How to integrate the modified \modelname~with the pre-trained model is also challenging. Injecting additional structure may destroy the representation ability of the pre-trained model. How to gently inject \modelname~into pre-trained models remains a difficult problem. \cite{jawahar2019does} proves that the bottom layer of PLMs pays more attention to words and syntactic information, and the higher layers pay more attention to semantic information.
Based on this conclusion, we disassembled the BERT main layer and verified the lack of differential information in different layers of BERT. By solving these problems, we can figure out which layers of BERT are missing differential information. Therefore, we use the robustness testing tool TextFlint as an experimental data set to study the above issues. First, TextFlint makes slight changes to each sampled example so that the sentence pairs have subtle differences. Second, we freeze the parameters of the BERT model (except the softmax classification output head) and adopt pre-trained contextualized word representations for the TextFlint task. This approach allows us to examine the extent to which syntax-related knowledge is stored in each layer of BERT and identify areas lacking this knowledge.

\begin{figure}[t]
\centering
\includegraphics[width=0.75\linewidth]{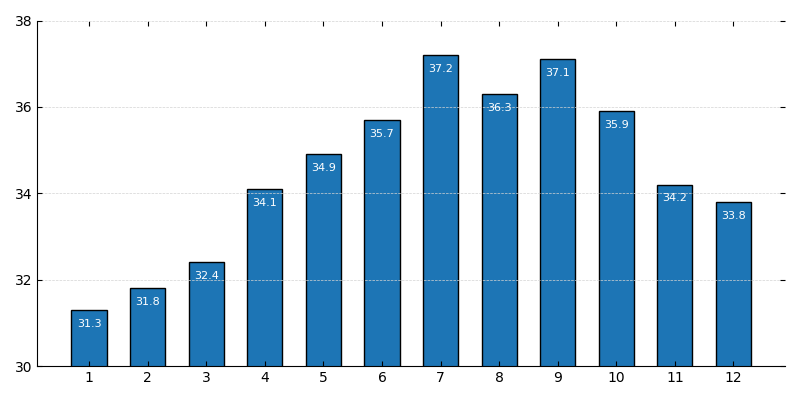}
\caption{Performance of each BERT layer on TextFlint transformed dataset.}
\label{fig:layer}
\end{figure}

Figure \ref{fig:layer} presents the performance of the BERT model layer-by-layer for difference awareness.  We leverage TextFlint to perform syntax structure transformations on the dataset, and the performance results are averaged over five different runs. A higher score indicates a stronger proficiency. From the figure, we observe that after freezing the layer parameters of BERT, the sensitivity to difference differs among the layers, with the middle and upper layers being more sensitive to difference than the lower layers. In summary, building upon the insights from the layer-by-layer analysis, we have identified a direction: incorporating \modelname~into the lower layers of BERT. 
In order to minimize the damage to the original pre-training process, we replace the multi-head attention in the first to third layers with \modelname~in the ratio of 50\%, 40\%, and 30\%.

\section{Experimental Settings}
\label{sec:exp}
\subsection{Datasets}

We conduct the experiments to test the performance of \modelname~on 10 large-scale publicly available sentence matching benchmark datasets. 
The GLUE benchmark~\cite{wang2018glue} is a widely used benchmark test suite in the field of NLP that encompasses various tasks such as sentence pair similarity detection and textual entailment\footnote{https://huggingface.co/datasets/glue}. 
We have conducted experiments on six sub-datasets of the GLUE benchmark: MRPC, QQP, STS-B, MNLI, RTE, and QNLI.
In addition to the GLUE benchmark, we also conduct experiments on four other popular datasets: SNLI~\cite{bowman2015large}, SICK~\cite{marelli2014sick}, TwitterURL~\cite{lan2017continuously} and Scitail~\cite{khot2018scitail}. 
The statistics of all 10 datasets are shown in Table~\ref{datasets-statistics}. 
Furthermore, to evaluate the robustness of the model, we also utilize the TextFlint~\cite{gui2021textflint} tool for robustness testing. 
TextFlint\footnote{https://www.textflint.io} is a multilingual robustness evaluation tool that tests model performance by making subtle modifications to the input samples.

\begin{table}[t]
\centering
\caption{The statistics of all 10 datasets.}
\label{datasets-statistics}
\resizebox{\linewidth}{!}{
\begin{tabular}{lcccccl}
    \toprule
    \textbf{Datasets} & \textbf{\#Train}& \textbf{\#Dev}& \textbf{\#Test}& \textbf{\#Label}& \textbf{Metrics}\\
    \midrule
    MRPC& 3669 & 409 & 1380 & 2 & Accuracy/F1\\
    QQP& 363871 & 1501 & 390965 & 2 & Accuracy/F1\\
    MNLI-m/mm& 392703 & 9816/9833 & 9797/9848 & 3& Accuracy \\
    QNLI& 104744 & 40432 & 5464 & 2 & Accuracy\\
    RTE& 2491 & 5462 & 3001 & 2 & Accuracy \\
    STS-B& 5749 & 1500 & 1379 & 2 & Pearson/Spearman corr\\
    \hline
    SNLI& 549367 & 9842 & 9824 & 3 & Accuracy\\
    SICK& 4439 & 495 & 4906 & 3 & Accuracy\\
    Scitail& 23596 & 1304 & 2126 & 2 & Accuracy\\
    TwitterURL& 42200 & 3000 & 9324 & 2 &Accuracy \\
    \bottomrule
\end{tabular}
}
\end{table}
 
\subsection{Baselines}
To evaluate the effectiveness of our proposed \modelname~in SSM, we mainly introduce BERT~\cite{devlin2018bert}, SemBERT~\cite{zhang2020semantics}, SyntaxBERT, UERBERT~\cite{xia2021using} and multiple other PLMs~\cite{devlin2018bert} for comparison. In addition, we also select several competitive models without pre-training as baselines, such as ESIM~\cite{chen2016enhanced}, Transformer~\cite{vaswani2017attention} , etc~\cite{hochreiter1997long,wang2017bilateral,tay2017compare}. 
In robustness experiments, we compare the performance of BERT on the robustness test datasets. 
For simplicity, the compared models are not described in detail here.

\begin{table*}[t]
\caption{The performance comparison of \modelname~with other methods.}
\label{citation-guide-outsideGlue}
\centering
 
\resizebox{\linewidth}{!}{
\begin{tabular}{lcccccccccccc}
\toprule
\textbf{Model} & \textbf{Pre-train}&\textbf{MRPC} & \textbf{QQP} & \textbf{STS-B} & \textbf{MNLI-m/mm}& \textbf{QNLI} & \textbf{RTE}  & \textbf{SNLI} & \textbf{Sci} & \textbf{SICK} & \textbf{Twi}  & \textbf{Avg}\\
\midrule
BiMPM &\XSolidBrush & 79.6 & 85.0 & - & 72.3/72.1 & 81.4 & 56.4 & - & - & - & - & -   \\
CAFE &\XSolidBrush & 82.4 & 88.0 & - & 78.7/77.9 & 81.5 & 56.8 & 88.5 & 83.3 & 72.3 & - & -  \\
ESIM &\XSolidBrush & 80.3 & 88.2 & - & 75.8/75.6 & 80.5 &  - & 88.0 & 70.6 & 71.8 & - & -   \\
Transformer &\XSolidBrush& 81.7 & 84.4 & 73.6 & 72.3/71.4 & 80.3 & 58.0 & 84.6 & 72.9 & 70.3 & 68.8 & 74.4 \\
\hline
BiLSTM+ELMo+Attnt&\Checkmark & 84.6 & 86.7 & 73.3 & 76.4/76.1 & 79.8 & 56.8 & 89.0 & 85.8 & 78.9 & 81.4 & 78.9  \\
OpenAI GPT &\Checkmark & 82.3 & 81.3 & 80.0 & 82.1/81.4 & 87.4 & 56.0 & 88.4 & 84.8 & 79.5 & 81.9 & 80.4  \\
UERBERT &\Checkmark & 88.3 & 90.5 & 85.1 & 84.2/83.5 & 90.6 & 67.1 & 90.8 & 92.2 & 87.8 & 86.2 & 86.0   \\
SemBERT & \Checkmark & 88.2 & 90.2 & 87.3 & 84.4/84.0 & 90.9 & 69.3 & 90.9 & 92.5 & 87.9 & 86.8 & 86.5  \\
SyntaxBERT &\Checkmark  & 89.2 & 89.6 & 88.1 & 84.9/84.6 & 91.1 & 68.9  & 91.0 & 92.7 & 88.7 & 87.3  & 86.3\\
DABERT &\Checkmark  & 89.1 &  91.3 & 88.2 & 84.9/84.7 & 91.4 & 69.5 & 91.3 & 93.6 & 88.6 & 87.5 & 86.7\\
\hline
BERT-Base&\Checkmark & 87.2 & 89.1 & 86.8 & 84.3/83.7 & 90.4 & 67.2 & 90.7 & 91.8 & 87.2 & 84.8 & 85.8\\
BERT-Base-\modelname~ &\Checkmark & \textbf{89.3} & \textbf{89.6} & \textbf{87.3} & \textbf{85.2/84.9} & \textbf{91.1} & \textbf{68.9} & \textbf{91.2} & \textbf{92.4} & \textbf{88.0} & \textbf{86.8} & \textbf{86.9}\\
\hline
BERT-Large&\Checkmark & 88.9 & 89.3 & 87.6 & 86.8/86.3 & 92.7 & 70.1 & 91.0 & 94.4 & 91.1 & 91.5 & 88.0\\
BERT-Large-\modelname~ &\Checkmark & \textbf{89.7} & \textbf{90.4} & \textbf{88.1} & \textbf{86.9/86.7} & \textbf{93.3} & \textbf{72.2} & \textbf{91.5} & \textbf{94.7} & \textbf{91.6} & \textbf{92.2} & \textbf{88.8}\\
\hline
RoBERTa-Base &\Checkmark& 89.3 & 89.6 & 87.4 & 86.3/86.2 & 92.2 & 73.6 & 90.8 & 92.3 & 87.9 & 85.9 & 87.6 \\
RoBERTa-Base-\modelname~  &\Checkmark & \textbf{89.8} & \textbf{91.1} & \textbf{88.4} & \textbf{87.5/87.4} & \textbf{93.7} & \textbf{82.3} & \textbf{91.2} & \textbf{93.2} & \textbf{89.6} & \textbf{87.7} & \textbf{89.2}\\
\hline
 RoBERTa-Large &\Checkmark& 89.4 & 89.7 & 90.2 & 89.5/89.3 & 92.7 & 83.8 & 91.2 & 94.3 & 91.2 & 91.9 & 90.3\\
 RoBERTa-Large-\modelname~ &\Checkmark  & \textbf{90.3} & \textbf{91.4} & \textbf{90.9} & \textbf{90.1/89.8} & \textbf{94.2} & \textbf{84.4} & \textbf{91.7} & \textbf{94.6} & \textbf{91.2}  & \textbf{92.2}& \textbf{90.9}\\
\bottomrule
\end{tabular}
}

\end{table*}

\begin{table}[t]
\caption{ Results of ablation experiment of various composition functions.}
\label{citation-guide-ablation}
\centering
\resizebox{\linewidth}{!}{
\begin{tabular}{lcc|cc|cc}
\toprule
\multirow{2}*{Model} &\multicolumn{2}{c}{QQP} &\multicolumn{2}{c}{QNLI} &\multicolumn{2}{c}{SNLI}\\ 
  \cmidrule(r){2-7}
   & \text{Dev} & \text{Test} & \text{Dev} & \text{Test} & \text{Dev} & \text{Test} \\
\midrule
\textbf{\modelname~} & \textbf{89.8} & \textbf{89.6}  & \textbf{92.2} & \textbf{91.1} & \textbf{92.2} & \textbf{91.1}\\
\text{$\tanh(\hat{E}) \displaystyle \odot sigmoid(N)$} & 89.7 & 89.3  & 92.4 & 91.2  & 92.4 & 91.2\\
\text{$\tanh(\textit{E}) \displaystyle \odot sigmoid(\hat{N})$} & 89.6 & 89.5 & 92.3 & 91.1 & 92.3 & 91.1\\
\midrule
\text{$\tanh(\textit{E}) \displaystyle \odot \tanh(N)$} & 86.5 & 85.2  & 87.3 & 85.8 & 87.3 & 85.8\\
\text{$\tanh(\textit{E}) \displaystyle \odot \arctan(N)$} & 85.1 & 84.6  & 86.4 & 84.3 & 86.4 & 84.3\\
\text{$sigmoid(\textit{E}) \displaystyle \odot \tanh(N)$} & 84.8 & 83.9  & 85.7 & 83.8 & 85.7 & 83.8\\
\text{$sigmoid(\textit{E}) \displaystyle \odot \arctan(N)$} & 86.2 & 85.0 & 87.4 & 85.6 & 87.4 & 85.6 \\
\text{$sigmoid(\textit{E}) \displaystyle \odot sigmoid(N)$} & 89.4 & 87.8 & 90.7 & 88.4  & 90.7 & 88.4\\
\bottomrule
\end{tabular}
}

\end{table}

\section{Results and Analysis}
\subsection{Model Performance}
 
To determine the efficacy of our method, we examine the effectiveness of aggregating \modelname~in 10 datasets, respectively.
Table \ref{citation-guide-outsideGlue} compares the performance of \modelname~and competing models across 10 datasets.
It is evident that the performance of non-pre-trained models is considerably inferior to that of pre-trained models. This is mainly because the pre-trained model has more data from learning corpus and powerful information extraction ability. When the backbone model is BERT-base or BERT-large, the average accuracy after integrating \modelname~is improved by 1.1\% and 0.8\%, respectively. The results show the effectiveness of our \modelname~Model on semantic matching tasks. In addition, our method outperforms RoBERTa-base by 1.6\% and RoBERTa-large by 0.6\%, respectively. which demonstrates that \modelname~can effectively capture the relationship between sentences from different aspects, so that more fine-grained and complex relationships can be exploited. 
These results demonstrate the advantages of combined attention modeling in mining semantics. 
Compared with previous work, our method shows very competitive performance levels in evaluating semantic similarity. In addition, the experimental results further verify the effectiveness of our method.

\begin{table*}[t]
\centering
\caption{\label{citation-guide-casestudy} The example sentence pairs of our cases. \textcolor{red}{Red} and \textcolor{blue}{Blue} are difference phrases. 
}

\begin{tabularx}{\textwidth}{l|X|X|X}
\toprule
\textbf{Case}  & \textbf{BERT} & \textbf{BERT-\modelname}  \\
\midrule
\text{S1:} Can \textcolor{red}{eat fruit for dinner} lead to weight loss?  & sim : 46.32\% & sim : 1.87\% \\
\text{S2:} Does \textcolor{blue}{ate dinner earlier} help with weight loss?  & label : 0 &label : 0 \\
\midrule
\text{S1:} How  \textcolor{red}{do girls} lose weight from \textcolor{red}{70 to 60} ?
 & sim : 72.66\% & sim : 12.06\% \\
\text{S2:} How \textcolor{blue}{should I} lose weight from \textcolor{blue}{60 to 50} ?
 & label : 1 & label : 0\\
\midrule
\text{S1:} What should I learn to be  a  \textcolor{red}{ hardware engineer}?
 & sim : 99.26\% \quad & sim : 18.63\%  \\
\text{S2:} What should I learn to be  a    \textcolor{blue}{software engineer}? & label : 1  &label : 0 \\
\bottomrule
\end{tabularx}
\end{table*}

\subsection{Ablation study}
To assess the individual impact of each component within our methodology, we have performed ablation experiments on the QQP, QNLI and SNLI datasets based on BERT. The experimental findings are shown in Table \ref{citation-guide-ablation}. In this study, we further examine the necessity of centering E and N, and the experimental results on the first two rows of three datasets. It has been discovered that centering matrix $E$ and $N$ does not help performance in most cases. Furthermore, it is seen that applying $Tanh$ on matrix $E$ and $Sigmoid$ on matrix $N$ outperforms other configurations for the proposed attention mechanism. This observation implicitly indicates the efficacy of the combinatorial attention.

\subsection{Robustness test performance}
\begin{figure}[t]
\centering
\includegraphics[width=1\linewidth]{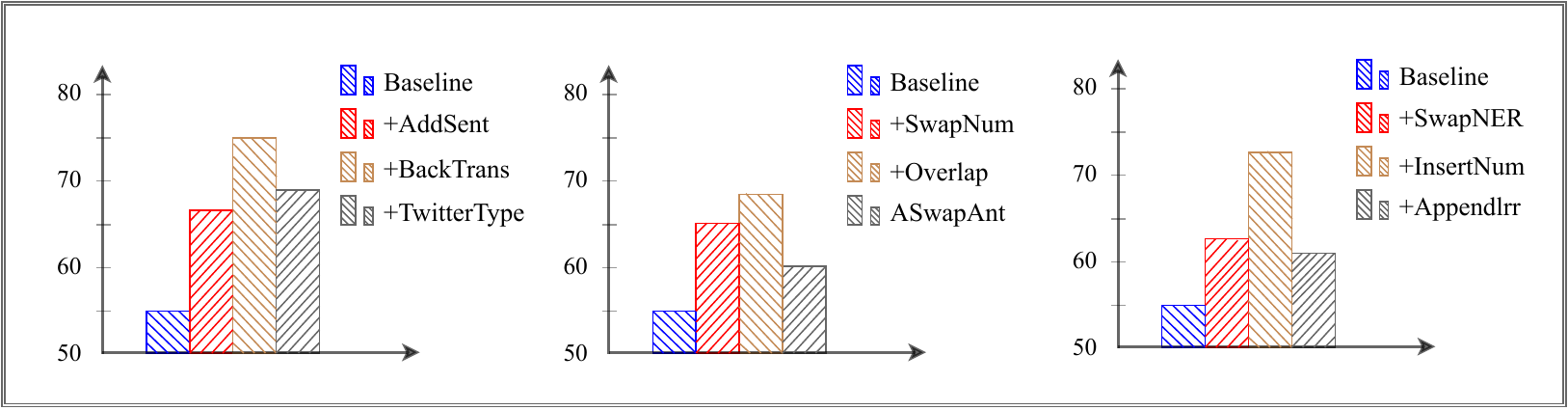}
\caption{The robustness experiment on the QQP and QNLI datasets based BERT.}
\label{fig:robust} 
\end{figure}

We conducted robustness tests on SNLI dataset. Figure \ref{fig:robust} lists the accuracy of \modelname~and BERT. We can observe that in SwapAnt our model outperforms BERT nearly 6\%, which indicates that \modelname~can better handle semantic contradictions caused by antonyms. And the model performance drops to 77.2\% on SwapNum transformation, while \modelname~outperforms BERT by nearly 5\% because it requires the model to capture subtle entity differences for correct linguistic inference. In other transformations, \modelname~still outperforms the baseline, which reflects its effectiveness.

\subsection{Case Study}
In order to intuitively understand how \modelname~works, we use the three cases in Table \ref{citation-guide-casestudy} for qualitative analysis. 
First, although S1 and S2 are literally similar in the first example, they express two completely different semantics due to the subtle difference the phrases bring to "eat fruit" and "eat early". The pre-trained language model BERT can identify semantic differences in case 1 and give correct predictions with the help of strong contextual representation capabilities. It is worth noting that the similarity of BERT's predicted sentence pairs is 46.32\%, while that of BERT-\modelname~is only 1.87\%. Second, in case 2, the sentence pairs "from 70 to 60" and "from 60 to 50" express different semantics, but they are primarily the result of numerical differences. Although BERT identified the correct label in case 1 by a small margin, in case 2, it was unable to capture numerically induced differences and gave wrong predictions because it requires the model to capture subtle numerical differences for correct language reasoning. Finally, our model made correct predictions in all of the above cases. Since \modelname~models sentence pairs from multiple perspectives, it can pay attention to the small differences in sentence pairs, and adaptively aggregate multi-source information in the alignment module to better identify the semantics within sentence pairs' differences.

\begin{figure}[t]
\centering
\includegraphics[width=0.95\linewidth]{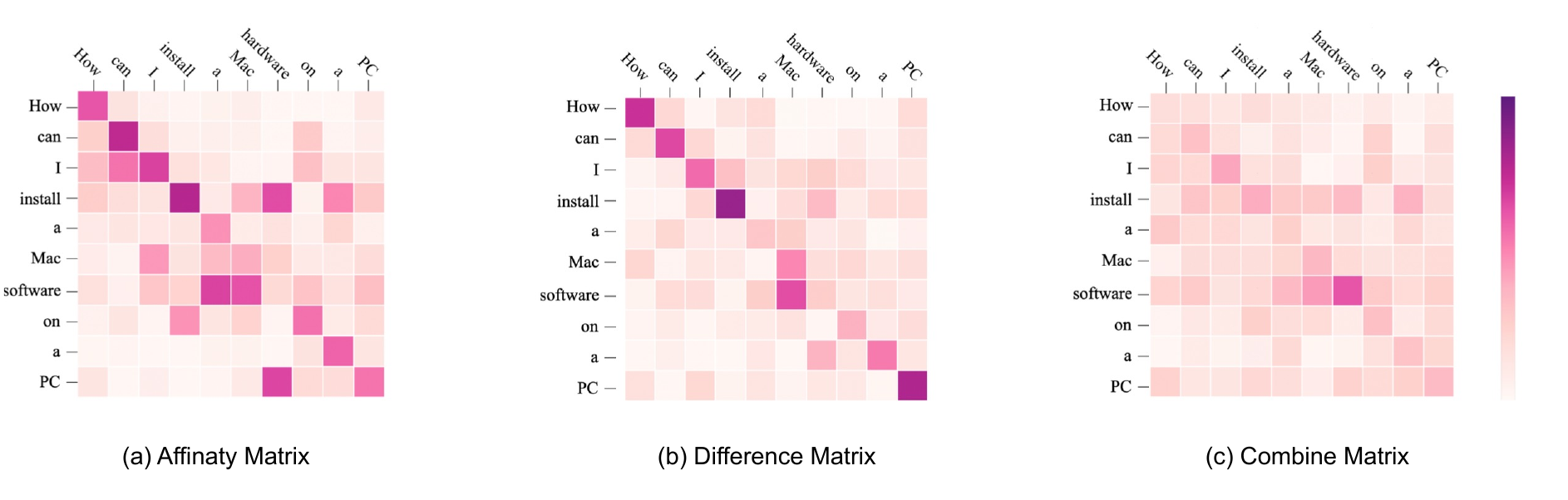}
\caption{Distribution of Affinaty Matrix (a), Difference Matrix (b), Combine Matrix (c)}
\label{fig:heatmap}
\end{figure}

\subsection{Attention Distribution}        
To visually demonstrate the impact of different attention functions inside multi-channel attention on the interactive alignment of sentence pairs, we show the weight distribution of three kinds of attention in Figure~\ref{fig:heatmap}. 
We can observe that the word-pair information in the sentence pairs concerned with different attention functions is inconsistent.
First, in Figure~\ref{fig:heatmap}(a), Dot attention can pay attention to the same words and semantically related words in sentence pairs, but it is heavily influenced by the same words in sentence pairs. It focuses too much on the shallow features of the same text and ignores the deep semantic association of the different words between "software" and "hardware". This shows that using Dot attention alone may lead to wrong predictions.
Secondly, in Figure~\ref{fig:heatmap}(b), it can be observed that Minus attention explicitly pays attention to the difference between "software" and "hardware", and its attention weight is the largest among all word pairs. This is because minus attention uses element-wise subtraction to compare the differences between sentence pairs. The greater the difference between word pairs, the greater their weight. Therefore, it can also be complementary to Dot attention.
Finally, in Figure~\ref{fig:heatmap}(c), the attention weights in combined attention focus on the same and different words, which shows that combined attention can both focus on the same part of the sentence pair and capture different parts, and this mechanism can capture both Affinity and dissimilarity of sentence pairs.
In summary, different attention focus on different word pairs in sentence pairs. Intuitively, our method can effectively combine the alignment relationships of multiple perspectives in sentence pairs to generate vectors that better describe the matching details of sentence pairs.

\section{Conclusion}
\label{sec:conclusion}
In this work, we propose a combination attention network based on transformer model for semantic sentence matching named \modelname. This model successfully captures the different information that is contained in pairs of words and integrates it into a model that has already been pre-trained.
The core of \modelname~lies in its dual-affinity module and compositional attention mechanism, which jointly capture the nuanced similarities and dissimilarities between sentence pairs. This unique capability enables \modelname~to discern subtle semantic differences that often evade traditional attention-based models.
The qualitative case study and attention distribution analysis provide clear insights into how \modelname~operates, revealing its ability to adaptively focus on relevant aspects of sentence pairs to enhance semantic understanding.
The results of our experiments on 10 publicly available datasets as well as a robustness dataset show that the consistent improvements across various metrics, especially the remarkable gains in robustness testing, underscore the effectiveness of our approach.
In future work, we will extend \modelname~to other NLP tasks and develop more sophisticated methods for integrating external knowledge into the model architecture.

\bibliography{main}
\end{document}